%% file: arxiv.tex
\pdfoutput=1

\documentclass[11pt]{article}
\usepackage{hyperref}
\usepackage{url}
\usepackage{amsmath}        
\usepackage{amsfonts}       
\usepackage{amssymb}        
\usepackage{mathrsfs}       
\usepackage{bm}             
\usepackage{amsthm}         
\usepackage{graphicx}
\usepackage{multirow}
\usepackage{subcaption}
\usepackage{graphicx}
\usepackage{microtype}
\usepackage{multirow}
\usepackage{footmisc}
\usepackage{soul}
\usepackage{arydshln}
\usepackage{multicol}
\usepackage{graphicx}
\usepackage{tcolorbox}
\usepackage{xcolor}
\tcbuselibrary{listings,breakable}
\usepackage{booktabs}
\usepackage{amssymb}
\usepackage{bbding}
\usepackage{pifont}
\usepackage{wasysym}
\usepackage{utfsym}
\usepackage{fontawesome}
\usepackage{listings}
\usepackage{xcolor}
\usepackage{multicol}
\usepackage{enumitem}
\usepackage{algpseudocode}
\usepackage{xspace}

\usepackage{times}
\usepackage{latexsym}
\usepackage{graphics}
\usepackage{graphicx}
\usepackage{color}
\usepackage{colortbl}
\usepackage{amssymb}
\usepackage{amsmath}
\usepackage{bbm}
\usepackage{booktabs}
\usepackage{arydshln} 
\usepackage{tcolorbox}
\usepackage{hyperref}
\usepackage{natbib}
\usepackage{xcolor}
\definecolor{myorange}{RGB}{2 142 2}
\usepackage{latex/acl}

\definecolor{deepgreen}{RGB}{0, 100, 0}
\usepackage{times}
\usepackage{latexsym}

\usepackage[T1]{fontenc}

\usepackage[utf8]{inputenc}

\usepackage{microtype}

\usepackage{inconsolata}

\usepackage{appendix}
\usepackage{tocbibind}

\usepackage{cleveref}
\crefformat{section}{\S#2#1#3}
\crefformat{subsection}{\S#2#1#3}
\crefformat{subsubsection}{\S#2#1#3}
\crefrangeformat{section}{\S\S#3#1#4 to~#5#2#6}
\crefmultiformat{section}{\S\S#2#1#3}{ and~#2#1#3}{, #2#1#3}{ and~#2#1#3}
\crefmultiformat{subsection}{\S\S#2#1#3}{ and~#2#1#3}{, #2#1#3}{ and~#2#1#3}
\Crefformat{figure}{#2Fig.~#1#3}
\Crefmultiformat{figure}{Figs.~#2#1#3}{ and~#2#1#3}{, #2#1#3}{ and~#2#1#3}
\Crefformat{table}{#2Tab.~#1#3}
\Crefmultiformat{table}{Tabs.~#2#1#3}{ and~#2#1#3}{, #2#1#3}{ and~#2#1#3}
\Crefformat{appendix}{Appx.~\S#2#1#3}
\crefmultiformat{appendix}{Appx.~\S#2#1#3}{ and~#2#1#3}{, #2#1#3}{ and~#2#1#3}
\crefformat{algorithm}{Alg.~#2#1#3}
\Crefformat{equation}{Eq.~#2#1#3}
\title{\raisebox{-0.3cm}{\includegraphics[width=1.0cm,height=1.0cm,keepaspectratio]{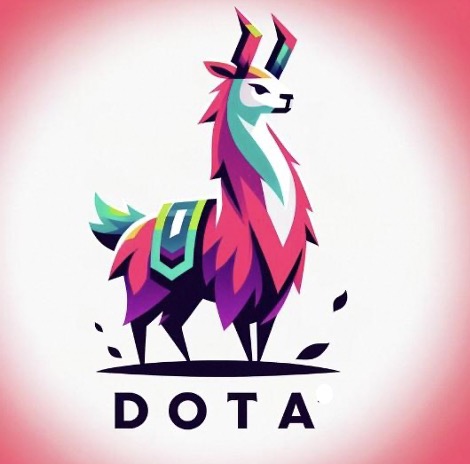}} 
DotaMath: Decomposition of Thought with Code Assistance and Self-correction for Mathematical Reasoning}

\author{Chengpeng Li\textsuperscript{12}\thanks{Work done during internship at Alibaba Group}, Guanting Dong\textsuperscript{2}$^*$, Mingfeng Xue\textsuperscript{2}$^*$, Ru Peng\textsuperscript{2}$^*$, Xiang Wang\textsuperscript{1}\\ 
\textbf{Dayiheng Liu\textsuperscript{2}\thanks{Corresponding author}} \\
\textsuperscript{1}University of Science and Technology of China \\
\textsuperscript{2}Alibaba Group.\\
{\{lichengpeng.lcp,liudayiheng.ldyh\}@alibaba-inc.com}\\
}

\begin{document}
\maketitle
\definecolor{deepgreen}{RGB}{0, 70, 0}
\definecolor{deepred}{RGB}{255, 0, 0}

\definecolor{backgreen}{RGB}{226, 240, 217}
\newcommand{\highg}{\cellcolor{backgreen}}
\begin{quote}
\textit{"Divide each difficulty into as many parts as is feasible and necessary to resolve it." -- René Descartes}
\end{quote} 
\begin{abstract}
Large language models (LLMs) have made impressive progress in handling simple math problems, yet they still struggle with more challenging and complex mathematical tasks.
In this paper, we introduce a series of LLMs that employs the \textbf{D}ecomposition \textbf{o}f \textbf{t}hought with 
code \textbf{a}ssistance and self-correction for \textbf{math}ematical reasoning, dubbed as \textbf{DotaMath}. DotaMath models tackle complex mathematical tasks by decomposing them into simpler logical subtasks, leveraging code to solve these subtasks, obtaining fine-grained feedback from the code interpreter, and engaging in self-reflection and correction. By annotating diverse interactive tool-use trajectories and employing query evolution on GSM8K and MATH datasets, we generate an instruction fine-tuning dataset called DotaMathQA with 574K query-response pairs. We train a series of base LLMs using imitation learning on DotaMathQA, resulting in DotaMath models that achieve the remarkable performance compared to open-source LLMs across various in-domain and out-of-domain benchmarks. Notably, DotaMath-deepseek-7B showcases an outstanding performance of 64.8\% on the competitive MATH dataset and 86.7\% on GSM8K. Besides, DotaMath-deepseek-7B maintains strong competitiveness on a series of in-domain and out-of-domain benchmarks (Avg. 80.1\%). Looking forward, we anticipate that the DotaMath paradigm will open new pathways for addressing intricate mathematical problems. Our code is
publicly available at \url{https://github.com/ChengpengLi1003/DotaMath}.

\end{abstract}

\input{latex/intro}

\section{Related Work}
\label{sec:relatedwork}
\paragraph{Mathematical Reasoning Techniques in LLMs.}
Early attempts to solve mathematical problems using large language models rely on basic computational expressions and results presented as few-shot examples~\citep{gpt3}.
Moreover, \citet{cot,kojima2022large} employ intermediate steps to enhance the capability of large language models in tackling complex mathematical reasoning tasks.
Building on this approach, \citet{pot,pal} introduce the use of code tools in the problem-solving process.
Subsequent researches explore various collaborative paradigms involving Chain-of-Thoughts and coding, which lead to significant improvements in the accuracy of solutions provided by large language models~\citep{yue2023mammoth,ToRA,MARIO-math,internMATH}. Different from them, 
we introduces DotaMath, a novel math problem-solving paradigm which decomposes mathematical problems into logical subtasks and utilizes code to address these tasks.
Our approach demonstrates superior performance across two in-domain and four out-of-domain math datasets.
\paragraph{Data Augmentation for Improved Math Reasoning.}
Beyond the exploration of paradigms, recent researches have increasingly focused on utilizing high-quality data to enhance the mathematical capabilities of large language models.
Some initiatives involve curating diverse collections of mathematical data to train specialized base models tailored specifically for mathematical tasks~\citep{llemma,OpenWebMath,MathPile,internMATH,deepseek}.
Other studies generate synthetic mathematical question-answer pairs by querying advanced large language models, such as GPTs~\citep{sun2023chatgpt}, Qwen~\citep{qwen}, and Mixtral~\citep{jiang2024mixtral}, to create Supervised Fine-Tuning (SFT) datasets~\citep{luo2023wizardmath,yu2023metamath,li2023query,mammoth2}.
In this paper, we presents a synthetic dataset aligned with the DotaMath paradigm, named DotaMathQA, which includes both single-turn-dialog form and multi-turn-dialog form that incorporate a self-correction process.
We demonstrate the effectiveness and generalizability of DotaMathQA across various backbone models and benchmarks.
\section{Method}
\begin{figure*}[t]
    \centering
    \small    \includegraphics[width=0.9\linewidth]{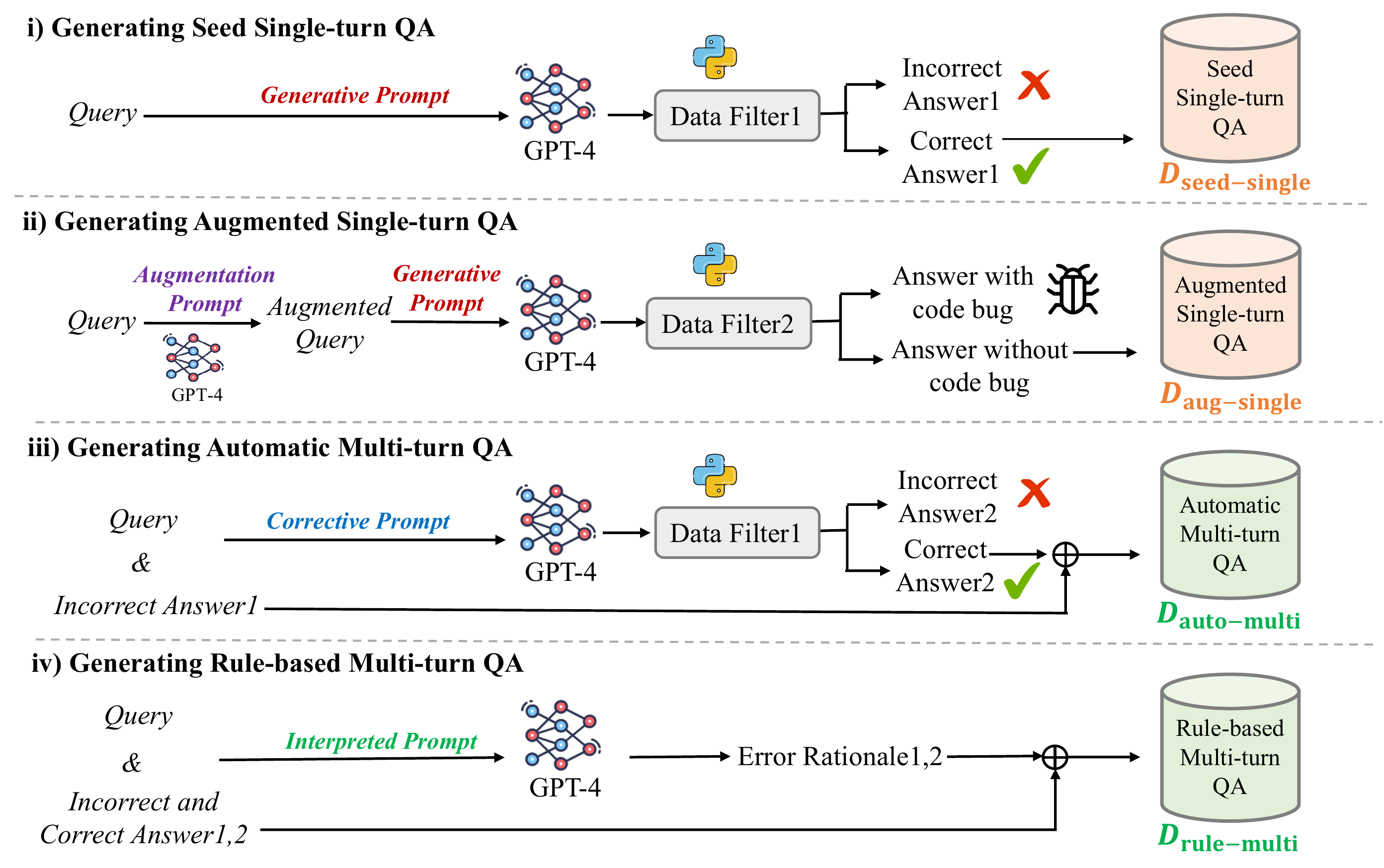}
    \caption{Dataset construction for DotaMathQA. The detailed desciption are in Section \ref{dataset_section}. All prompts used are listed in Appendix \ref{app:prompt}. Data Filter1 means retaining correct instances by comparing the results of the generated code from the Python interpreter with the ground truth answers in the dataset. Data Filter2 means filtering out instances where the generated code results in execution errors.
    }\vspace{-5pt}
    \label{data_produce}
\end{figure*}

In this section, we first introduce how DotaMath performs mathematical reasoning through interaction with Python interpreter (\Cref{sec: dotaformatting} \& \Cref{data_format}). Next, we introduce the pipline of using GPT-4 for data annotation to synthesize the instruction-tuning dataset, DotaMathQA.\footnote{In this paper, all data generated by GPT-4 are derived from the gpt-4-turbo-2024-04-09 api.} (\Cref{dataset_section} \& \Cref{data_produce}). Finally, we discuss the process of supervised fine-tuning a range of foundational LLMs on the DotaMathQA dataset (\Cref{sec:sft}).

\subsection{Inference Procedure}
\label{sec: dotaformatting}



Motivated by a series of efforts that integrate the Python interpreter's output as supervision~\citep{le2022coderl,chen2023teaching,qiao2023making,dong2024self}, DotaMath solves mathematical problems through several operations, including task decomposition, writing Python programs, invoking the Python interpreter and self-correction (Figure~\ref{data_format}). For a given problem $q$ and system prompt $p$ in Appendix~\ref{systemprompt}, DotaMath($\mathcal{M}$) initially decompose it into some sub-tasks, yielding $d_1 = d_1^1 \oplus d_1^2 \oplus d_1^3$, where $\oplus$ means concatenation. 
\begin{equation}
    d_1 \sim \mathbb{P}_{\mathcal{M}}(\cdot \mid p \oplus q ),
\end{equation}
where $\mathrm{P}_{\mathcal{M}}$ means the probability distribution of a LLM $\mathcal{M}$.
Subsequently, DotaMath generates a Python code segment $c_1$ to address these sub-tasks, presenting the reasoning process of each sub-task internally via substituting the intermediate results of sub-tasks with variables in the code. In line with with~\citep{ToRA}, the Python code utilizes a specific start token \verb|"```Python"| and a distinct end token \verb|"```output"| to determine when to cease generation for invoking the Python interpreter:
\begin{equation}
    c_1 \sim \mathbb{P}_{\mathcal{M}}(\cdot \mid p \oplus q \oplus d_1).
\end{equation}
The execution result $r_1 = r_1^1 \oplus r_1^2 \oplus r_1^3$ (including results of all subtasks) obtained via the Python interpreter is fed back to DotaMath for further analysis. If the problem is resolved, DotaMath will generate the final result $a$ and place the answer within \verb|"\boxed{}"| for evaluation:
\begin{equation}
    a \sim \mathbb{P}_{\mathcal{M}}(\cdot \mid p \oplus q \oplus d_1 \oplus c_1 \oplus r_1) .
\end{equation}
This type of data is referred to as single-turn QA. Otherwise, DotaMath reflects on the previous decomposition and code based on the results of all sub-tasks, providing an explanation for any errors $e_1$:
\begin{equation}
    e_1 \sim \mathbb{P}_{\mathcal{M}}(\cdot \mid p \oplus q \oplus d_1 \oplus c_1 \oplus r_1).
\end{equation}
Subsequently, the process of task decomposition, code generation, and invoking the Python interpreter is repeated until the problem is resolved or a predetermined maximum number of tool invocations is reached. This iterative process is referred to as self-correction, and the corresponding data is termed multi-turn QA. Overall, the interaction trajectory between the model and the Python interpreter can be summarized as follows:
\begin{equation}
    \tau = d_1\oplus c_1 \oplus r_1 \oplus e_1 \cdots d_n\oplus c_n \oplus r_n \oplus a.
    \label{trajectory}
\end{equation}

\begin{table*}[ht]
\centering
\small
\renewcommand{\arraystretch}{1.2}
\resizebox{0.98\textwidth}{!}{
\begin{tabular}{lcccccl}
\hline
\toprule
\textbf{Datasets} & \textbf{Size} & \textbf{LLM Used?} & \textbf{Use Tool?} &\textbf{Data Sources} \\
\hline
RFT~\citep{yuan2023scaling} & 100K & LLaMA-2 & $\times$  & GSM8K \\
Open-Platypus~\citep{Platypus} & 25K & GPT-4 &  $\times$ & 11 datasets with MATH \\
WizardMath~\citep{luo2023wizardmath} & >50K & ChatGPT & $\times$  & MATH \& GSM8K \\
Lila~\citep{lila} & 134K & - &  \checkmark & 20 datasets with MATH \& GSM8K \\
MathInstruct~\citep{yue2023mammoth}& 260K & GPT-4 &  \checkmark & 14 datasets with GPT-4 \\
ToRA~\citep{ToRA} & 69K & GPT-4 & \checkmark &  MATH \& GSM8K \\
MetaMath~\citep{yu2023metamath} & 395K & GPT-4 & \textbf{$\times$} &  MATH \& GSM8K \\
MuggleMATH~\citep{li2023query} & 600K & GPT-4 & \textbf{$\times$} &  MATH \& GSM8K \\
DotaMathQA& 574K & GPT-4 & \checkmark &  MATH \& GSM8K   \\
\bottomrule
\end{tabular}
}
\caption{Dataset Statistics and Comparison.}
\vspace{-5pt}
\label{Dataset Comparision}
\end{table*}

\subsection{DotaMathQA Dataset Construction}
\label{dataset_section}

As illustrated in Figure \ref{data_produce}, our DotaMathQA dataset can be divided into four parts: $\{\mathcal{D}_\textbf{seed-single}, \mathcal{D}_\textbf{aug-single}, \mathcal{D}_\textbf{auto-multi},\mathcal{D}_\textbf{rule-multi}\} \in \mathcal{D}_\textbf{DotaMathQA}$. As illustrated in Figure~\ref{data_produce}, queries for $\mathcal{D}_\textbf{seed-single}$, $\mathcal{D}_\textbf{auto-multi}$, and $\mathcal{D}_\textbf{rule-multi}$ originate from two popular mathematical reasoning datasets, GSM8K and MATH. Additionally, queries for $\mathcal{D}_\textbf{aug-single}$ come from the augmented queries of these datasets. The single-round data $\mathcal{D}_\textbf{single} = \{\mathcal{D}_\textbf{seed-single},\mathcal{D}_\textbf{aug-single}\}$ involves one-time tool usage (Python interpreter), excluding the self-correction process. In contrast, the multi-round data $\mathcal{D}_\textbf{multi} = \{\mathcal{D}_\textbf{auto-multi},\mathcal{D}_\textbf{rule-multi}\}$ encompasses multiple tool invocations, including the self-correction process.



\paragraph{$\mathcal{D}_\textbf{seed-single}$ Construction.} 
Leveraging the powerful instruction-following capability of GPT-4, we prompt it to generate solutions in the desired DotaMath format for queries from the GSM8K and MATH training datasets. This is achieved by manually writing a single query-response demonstration derived from the MATH training set to guide GPT-4 in producing the corresponding data format, as outlined in the \textbf{generative prompt} in \textbf{Appendix} \ref{generativeprompt}. We conduct nucleus sampling with a temperature of 0.5 and top-p of 1.0, generating four responses per query. Subsequently, we pass all the generated responses to the Python interpreter. We employ both rules and manual inspection to verify whether the answers match the reference answers in the original GSM8K and MATH datasets. If they match, we place them in $\mathcal{D}_\textbf{seed-single}$. If the solutions generated by GPT-4 are incorrect, we revert to using the \textbf{generative prompt} with a temperature of 0.7 and top-p of 1.0 to sample 10 more responses per query. This re-annotation process is repeated up to five times.

\paragraph{$\mathcal{D}_\textbf{aug-single}$ Construction.}
Utilizing GPT-4 to bootstrap queries in the training dataset for diversification significantly enhances the in-domain reasoning capabilities of large models~\citep{luo2023wizardmath,yu2023metamath,li2023query}.  
Similar to~\citeauthor{li2023query}, we modify mathematical problems in GSM8K and MATH for query augmentation using the augmentation prompt in Appendix \ref{app:prompt}. 
Then, we annotate the augmented queries them using \textbf{generative prompt} with GPT-4. Given the absence of standard answers for augmented queries, we only filter out responses that contain python execution bugs.

\paragraph{$\mathcal{D}_\textbf{auto-multi}$ Construction.}
\label{auto-multi}
For incorrect solutions generated by GPT-4, we use the \textbf{corrective prompt} in Appendix \ref{Correctionprompt} to instruct GPT-4 to correct them. The new solution is then sent to the Python interpreter. If the self-correction is successful, we concatenate the incorrect and corrected solutions to create error-correcting data, adding this data to $\mathcal{D}_\textbf{auto-multi}$. We fine-tune several DeepSeekMATH-Base models using subsets of $\mathcal{D}_\textbf{seed-single}$ and then use them to annotate the training sets of GSM8K and MATH. This process aims to enrich the diversity of queries and responses in the error-correcting data, enabling the model to identify and correct a wider range of error patterns.


\paragraph{$\mathcal{D}_\textbf{rule-multi}$ Construction.} The efficiency of obtaining error-correcting data through \ref{auto-multi} has been observed to be relatively low due to GPT-4's limited success rate in correcting incorrect responses. To address the issue of low efficiency in obtaining self-correction data, a new method for constructing self-correction data has been designed. For queries that possess both correct and incorrect responses, GPT-4 is directed to analyze the incorrect solution based on the correct one and explain the reasons for the error. The incorrect solution, the explanation of its error, a linking sentence \verb|"let's correct the solution"|, and the correct solution are then concatenated together to form new self-correction data.

Table~\ref{Dataset Comparision} compares DotaMathQA with several recently introduced mathematical datasets. Overall, $D_\textbf{seed-single}$, $D_\textbf{auto-multi}$ and $D_\textbf{rule-multi}$ensure that 99.3\% of GSM8K queries and 96\% of MATH queries have at least one correct solution or a solution that has undergone correction. The sizes of $D_\textbf{seed-single}$, $D_\textbf{auto-multi}$,$D_\textbf{rule-multi}$ and $D_\textbf{aug-single}$ are 80K, 2K, 10K, and 482K, respectively.

\subsection{Supervised Fine-tuning}
\label{sec:sft}

We describe $\mathcal{D}_\textbf{{DotaMathQA}}$ as $\mathcal{D}_\textbf{DotaMathQA} = \{ (q_i, \tau_i) \}_{i}$, where $q_i$ represents a math question and $\tau_i$ indicates an interaction trajectory of natural language and tools in Eq.\ref{trajectory}. We apply supervised fine-tuning on a series of base models, including Llama2-7B~\citep{llama2}, LLaMA3-8B~\citep{llama3}, Llemma-7B~\citep{llemma}, and DeepSeekMath-Base-7B~\citep{deepseek}, resulting in a series of DotaMath-LLMs. For a LLM (parameterized by $\boldsymbol{\theta}$), the  optimization objective is to maximize the log likelihood of the reasoning trajectory conditioned on the question,
\begin{equation}
  \mathcal{L}(\boldsymbol{\theta})=\sum_{(q_i, \tau_i) \in \mathcal{D}_{\text {DotaMathQA }}} \log \mathbb{P}(\tau_i \mid \hat{q_i} ; \boldsymbol{\theta}),
\end{equation}

where $\hat{q_i}$ represents the content of $q_i$ equipped with a system prompt in Appendix~\ref{systemprompt}.


\begin{table*}[ht]
\centering
\renewcommand{\arraystretch}{1.2}
\resizebox{0.98\textwidth}{!}{
\begin{tabular}{@{}lccccccccc@{}}
\toprule
\textbf{Model}  &\textbf{Size} &\textbf{Use Tool?} & \multicolumn{2}{c}{\textbf{In-domain}} & \multicolumn{4}{c}{\textbf{Out-of-domain}}\\
\cmidrule(lr){4-5} \cmidrule(lr){6-9} 
   &   &      & GSM8K & MATH & Mathematics & SVAMP & TabMWP & ASDiv & Average\\
\midrule
\multicolumn{8}{c}{Proprietary Model} \\
\midrule
Claude-3 Opus~\citep{claude-3} & - &\color{red}\ding{53}  & 95.0 & 60.1  & -  & -  & - & - & - \\
GPT-4(original version)~\citep{gpt4} & - &\color{red}\ding{53}  & 92.0 & 42.5 & - & 93.1 & 67.1 & 91.3 & -  \\
GPT-4 Code Interpreter~\citep{gpt4-code-early} & - &\color{deepgreen}\checkmark & 97.0 & 69.7 & - & - & - & -& - \\
GPT-4 (PAL)~\citep{ToRA}& - &\color{deepgreen}\checkmark    & 94.2 & 51.8 &  & 94.8 & 95.9 & 92.6  & - \\

\bottomrule
\multicolumn{8}{c}{Open-Source Model} \\
\midrule

MetaMATH~\citep{yu2023metamath}    & 70B   &\color{red}\ding{53}      &82.3  &26.6    &- &   85.8& 63.4& 84.0  &-  \\
MuggleMATH~\citep{li2023query} & 70B   &\color{red}\ding{53}    &82.7  &  36.3     & 34.6 &83.4 &59.7 &  81.8  &   57.8  \\
MAmmoTH~\citep{yue2023mammoth}     & 70B     &\color{deepgreen}\checkmark  &76.9  &41.8    &55.6   & 82.4  &38.2 &70.2    &48.0   \\
ToRA~\citep{ToRA}      & 70B     &\color{deepgreen}\checkmark  &84.3 &49.7 &72.6 &82.7 &74.0 &86.8   &70.5    \\
MathGenieLM~\citep{lu2024mathgenie} & 70B &  \color{deepgreen}\checkmark&\textbf{88.4} &51.2 &76.0 & 87.7 & - & - & -\\
\hline
ToRA~\citep{ToRA}      & 7B     &\color{deepgreen}\checkmark  &68.8 &40.1 &58.3 &68.2& 42.4& 73.9   &49.0    \\
MathGenieLM~\citep{lu2024mathgenie} & 7B &  \color{deepgreen}\checkmark&71.7 &33.0 &65.0 & 78.5 & - & - & -\\
\textbf{DotaMath-LLaMA2-7B}     & 7B  &\color{deepgreen}\checkmark     & 79.6 & 50.1       & 65.0 & 84.2 &  64.1  &82.3    &64.7 \\
MAmmoTH2-8B & 8B &\color{red}\ding{53}&70.4 & 35.8 &-&-&-&- & -\\
\textbf{DotaMath-LLaMA3-8B  }   & 8B  &\color{deepgreen}\checkmark     & 84.2 & 58.9      & 74.2 &\underline{88.3} & 70.4 & 85.1        &71.1  \\
ToRA-llema~\citep{ToRA} & 7B &\color{deepgreen}\checkmark   &74.8	&49.5 & -& 76.0&	63.5&	82.3 & -\\
MARIO-llema-7B~\citep{MARIO-math}  & 7B        &\color{deepgreen}\checkmark   & 70.1&	46.3   & -   & -   & -   & -   & -       \\
\textbf{DotaMath-llemma-7B}     & 7B   &\color{deepgreen}\checkmark    & 81.2 & \underline{57.7}     & 76.8 & 87.8 &  77.7 & 85.1     &74.3 \\

DeepSeek-MATH-Instruct~\citep{deepseek}&7B      &\color{deepgreen}\checkmark         & 83.7 & 57.4\      & \textbf{82.8} &85.7 &  \underline{79.7}  &86.6 &\underline{75.7}  \\
KPMath-Plus-deepseek~\citep{keypoint} &7B &\color{red}\ding{53}  &83.9 &48.8 &-& 81.5& 78.7 &\textbf{88.9} & - \\
MARIO-deepseek~\citep{MARIO-math} &7B &\color{deepgreen}\checkmark  &78.4 &56.1 &-&-&-&- & -\\
\textbf{DotaMath-deepseek-7B}   & 7B &\color{deepgreen}\checkmark        & \underline{86.7} & \textbf{64.8} & \underline{79.1}&\textbf{89.5}&\textbf{84.2}&\underline{88.5}     &\textbf{80.1}  \\

\bottomrule
\end{tabular}
}
\caption{Comparison of DotaMath with open-source and proprietary source LLMs on in-domain and out-of-domain benchmarks. The base model of open-source LLMs above DotaMath-LLaMA2-7B are LLaMA2. The average score is derived from a weighted average of scores across benchmarks, with weights the proportion of questions in each benchmark. The best results are highlighted in bold, and the second-best results are underlined.
}
\label{mainresult_table}
\end{table*}

\section{Experiments}


\subsection{Experimental Setup}

\paragraph{Datasets.}
The models are evaluated on two in-domain datasets, GSM8K~\citep{gsm8k} and MATH~\citep{MATH}, as wel as four out-of-domain datasets: Mathematics~\citep{mathimatics}, SVAMP~\citep{svamp}, TabMWP~\citep{tabmvp}, and ASDiv~\citep{asdiv}. These six datasets encompass a wide range of difficulty levels, from grade school math word problems to challenging competition mathematics problems, covering multiple domains including Algebra, Number Theory, Counting and Probability, Geometry, and Precalculus. They include various types of questions, such as tabular-based, free-form, and multiple-choice, offering a comprehensive assessment of the model's mathematical reasoning capabilities.

\paragraph{Metrics.} We use the accuracy of predicted answers to evaluate LLMs. Following~\citep{metric}, we round numbers and parsed expressions using \textit{sympy2}. 


\subsection{Baselines}
We benchmark our models with following proprietary Models and Open-Source Models:
\begin{itemize}
    \item \textbf{Proprietary Models}: Claude-3~\citep{claude-3}, GPT-3.5~\citep{gpt3}, GPT-4~\citep{gpt4}, etc.
    \item  \textbf{Open-Source Models}: WizardMATH~\citep{luo2023wizardmath}, MetaMATH~\citep{yu2023metamath}, MuggleMATH~\citep{li2023query}, RFT~\citep{yuan2023scaling}, MATHCoder~\citep{mathcoder}, ToRA~\citep{ToRA}, MARIO~\citep{MARIO-math}, etc.
\end{itemize}
For space saving, only part results are listed in Table \ref{mainresult_table}. More results can be found in Table \ref{other_mainresult_table}.

\subsection{Main Results}

Table~\ref{mainresult_table} compares DotaMath with a range of state of art mathematical methods across in-domain and out-of-domain benchmarks. We can draw several conclusions: 

(1)
On the elementary mathematical task GSM8K, most state-of-the-art 70B open-source models achieve a performance of over 80, regardless of tool usage. This indicates that the tool-based paradigm does not offer a significant advantage over Chain-of-Thought (COT) methods in simple mathematical reasoning tasks. However, DotaMath, with a size of just 7B, surpasses most of the 70B open-source models, demonstrating strong competitiveness.

\begin{table}[t]
\centering
\renewcommand{\arraystretch}{1.2}
\resizebox{0.48\textwidth}{!}{
\begin{tabular}{lcc}
\toprule
Traning Set& GSM8K & MATH \\
\midrule
$\mathcal{D}_\textbf{DotaMathQA-w/o-aug}$  & 82.4 & 59.0 \\
$\mathcal{D}_\textbf{DotaMathQA-w/o-aug-w/o-dot}$ & 80.0($\downarrow$ 3.0\%) & 55.5($\downarrow$ 6.0\%) \\
$\mathcal{D}_\textbf{DotaMathQA-w/o-aug-w/o-inter}$ & 81.5($\downarrow$ 1.1\%) &57.4($\downarrow$ 2.7\%)  \\
$\mathcal{D}_\textbf{DotaMathQA-w/o-aug-w/o-multi}$ & 81.3($\downarrow$ 1.3\%) & 56.7($\downarrow$ 3.9\%) \\

\bottomrule
\end{tabular}
}
\caption{The ablation study of decomposition of thought, display intermediate process and self-correction.}
\vspace{-5pt}
\label{tab:format ablation}
\end{table}

(2)
On the competition-level mathematical task MATH, models utilizing tools significantly outperform those that do not, emphasizing the necessity of the tool-based paradigm for complex mathematical tasks. DotaMath substantially outperforms all open-source models and even exceeds the strongest proprietary model, Claude-3 Opus.

(3)
The DotaMath series also demonstrates the best performance on untrained Out-of-Domain datasets, indicating our model's strong generalization capabilities and comprehensive mathematical reasoning abilities. On average, DotaMath-deepseek outperforms the previous best open-source SFT model, DeepSeek-MATH-Instruct, by 4.4 points..

(4)
On in-domain benchmarks, DotaMath-LLaMA2-7B, DotaMath-llemma-7B, DotaMath-LLaMA3-7B, and DotaMath-deepseek-7B exhibit incremental performance improvements. These differences are likely attributable to the quantity and quality of math-related data used in their pre-training or continual pre-training.



\subsection{Ablation Study}
To verify whether our designs enhance the mathematical reasoning capabilities of models, we performed ablation studies on different parts of the data. All results are based on fine-tuning the DeepSeekMath-Base model. In summary, we conducted ablation studies on two components: data format and data augmentation. The data format ablation includes three aspects: decomposition of thought, intermediate process display, and self-correction.

\begin{figure}[t]
    \centering
    \resizebox{0.47\textwidth}{!}{
    \includegraphics{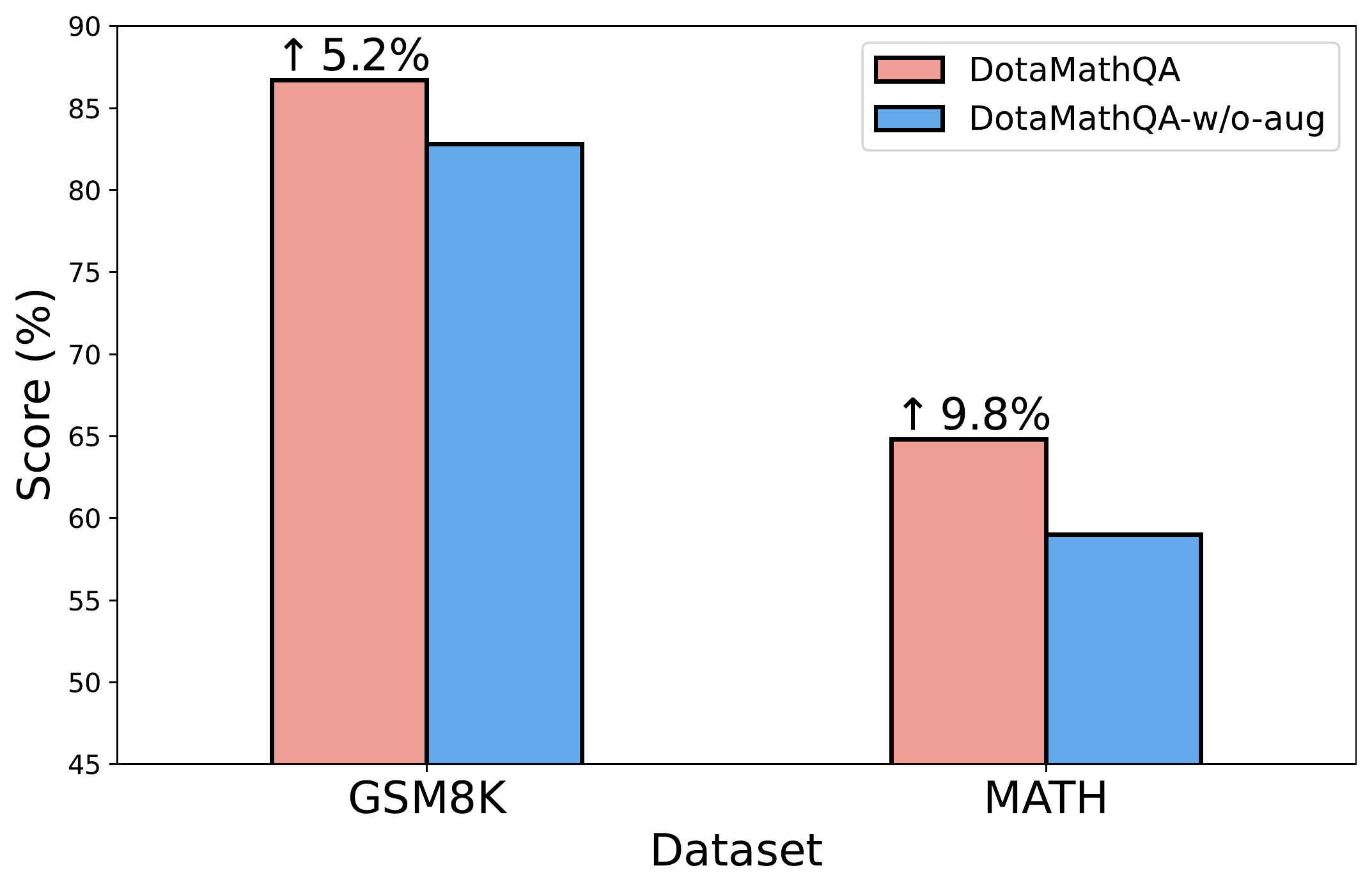}
    }
    \vspace{-10pt}
    \caption{The comparison of DeepSeekMath-Base fine-tuned with $\mathcal{D}_\textbf{DotaMathQA}$ and $\mathcal{D}_\textbf{DotaMathQA-w/o-aug}$ on GSM8K and MATH.}
    \label{fig:enter-label}
    \vspace{-5pt}
\end{figure}

\begin{table*}[t]
\centering
\renewcommand{\arraystretch}{1.2}
\resizebox{0.95\textwidth}{!}{
\begin{tabular}{@{}lccccc@{}}
\toprule
Model & Level 1 & Level 2 & Level 3 & Level 4 & Level 5 \\
\midrule
$\mathcal{D}_\textbf{DotaMathQA}$ & 91.3 ($\uparrow$ 4.5\%) & 80.5 ($\uparrow$ 6.3\%) & 74.7 ($\uparrow$ 11.8\%) & 59.3 ($\uparrow$ 8.0\%) & \highg{41.9 ($\uparrow$ 18.7\%)} \\
$\mathcal{D}_\textbf{DotaMathQA-w/o-aug}$ & 87.4 & 75.7 & 66.8 & 54.9 & 35.3 \\
$\mathcal{D}_\textbf{DotaMathQA-w/o-aug-w/o-dot}$ & 85.1 ($\downarrow$ 2.6\%) & 72.1 ($\downarrow$ 4.8\%) & 65.4 ($\downarrow$ 2.1\%) & 49.6 ($\downarrow$ 9.7\%) & 31.6 \highg{($\downarrow$ 10.5\%)} \\
$\mathcal{D}_\textbf{DotaMathQA-w/o-aug-w/o-inter}$ & 85.8 ($\downarrow$ 1.8\%) & 73.3 ($\downarrow$ 3.2\%) & 66.6 ($\downarrow$ 0.3\%) & 53.7 ($\downarrow$ 2.2\%) & 33.3 \highg{($\downarrow$ 5.7\%)} \\
$\mathcal{D}_\textbf{DotaMathQA-w/o-aug-w/o-multi}$ & 86.7 ($\downarrow$ 0.8\%) & 73.9 ($\downarrow$ 2.4\%) & 64.3 ($\downarrow$ 3.7\%) & 52.4 ($\downarrow$ 4.6\%) & \highg{32.8 ($\downarrow$ 7.1\%)} \\
\bottomrule
\end{tabular}
}
\caption{Sub-levels performance of different models on MATH. Level5 is hardest.The most significant change appears in \colorbox{backgreen}{green}. The number of questions from Level 1 to Level 5 are 437, 894, 1131, 1214, and 1324, respectively.}
\label{tab:levels}
\end{table*}

\paragraph{Ablation on Data Format.}
Given that augmented data significantly increases in volume compared to the original dataset and cannot guarantee accuracy, our ablation studies of data format are conducted exclusively on the training sets of GSM8K and MATH and we name it $\mathcal{D}_\textbf{DotaMathQA-w/o-aug}=\{\mathcal{D}_\textbf{seed-single},\mathcal{D}_\textbf{multi}\}$. To verify the role of decomposition of thought, we eliminate the rationale that decomposes the question into sub-tasks from the responses in the $\mathcal{D}_\textbf{DotaMathQA-w/o-aug}$ dataset and get $\mathcal{D}_\textbf{DotaMathQA-w/o-aug-w/o-dot}$. From Table \ref{tab:format ablation}, we observe a decrease in accuracy of 3.0\% on GSM8K and 6.0\% on MATH, underscoring the effectiveness of decomposition of thought, particularly for the more challenging MATH dataset. Similarly, we remove all print statements from the code in the dataset responses except those printing the final result, and using a Python interpreter, we execute the modified code to obtain new data $\mathcal{D}_\textbf{DotaMathQA-w/o-aug-w/o-inter}$. The corresponding performance decrease is 1.1\% on GSM8K and 2.7\% on MATH, illustrating the effectiveness of displaying intermediate processes.  To examine the role of error correction, we remove $\mathcal{D}_\textbf{multi}$ from $\mathcal{D}_\textbf{DotaMathQA-w/o-aug}$ to get $\mathcal{D}_\textbf{DotaMathQA-w/o-aug-w/o-multi}$. Following this removal, the model's performance decrease by 1.3\% on GSM8K and 3.9\% on MATH, indicating that self-correction enhances the model's ability to solve complex tasks.
    

\paragraph{Ablation on Data Augmentation.}
To analyze the impact of data augmentation, we separately fine-tune the DeepSeekMath-Base model on $\mathcal{D}_\textbf{DotaMathQA}$  and $\mathcal{D}_\textbf{DotaMathQA-w/o-aug}$. On GSM8K, data augmentation increases the accuracy from 82.4 to 86.7, marking a 4.7\% improvement. On MATH, data augmentation boost the performance from 59.0 to 64.8, achieving a 9.8\% increase. These performance improvements further demonstrate the significant role of query augmentation in enhancing the model's mathematical reasoning capabilities. The disparity in performance gains between GSM8K and MATH could be aligned to the finding~\citep{li2023query} that the model fine-tuned without augmented data already achieves a high accuracy on GSM8K, making further enhancements more difficult.

\subsection{Analysis}

\begin{table}[t]
\centering
\resizebox{0.45\textwidth}{!}{
\begin{tabular}{lcc}
\toprule
Training Set& GSM8K & MATH \\
\midrule
$\mathcal{D}_\textbf{seed-single}$ &  81.3 & 56.7 \\
$\mathcal{D}_\textbf{seed-single}$+$\mathcal{D}_\textbf{auto-multi}$(2K) &  81.8 & 58.4 \\
$\mathcal{D}_\textbf{seed-single}$+$\mathcal{D}_\textbf{rule-multi}$(10K) & 82.3 & 58.6 \\
$\mathcal{D}_\textbf{seed-single}$+$\mathcal{D}_\textbf{rule-multi}$(2K) & 81.6 & 57.5 \\
$\mathcal{D}_\textbf{seed-single}$+$\mathcal{D}_\textbf{multi}$ & 82.8 & 59.0 \\
\bottomrule
\end{tabular}
}
\caption{Effectiveness of different self-correction data.}
\label{gpt4vsrule}
\end{table}


\paragraph{The Effectiveness of Data Augmentation in MATH Sub-levels.}

For data augmentation, the performance gains at different levels are roughly positively correlated with question difficulty. This indicates that data augmentation is more effective for more difficult questions. At the most difficult Level 5, data augmentation increases the model's performance from 35.5 to 41.9, achieving an impressive 18.7\% improvement. For decomposition of thought, Intermediate Process Display, and self-correction, the most significant changes also occur with Level 5 questions. This indicates that such designs are effective in enhancing the model's reasoning capabilities for complex problems.

\begin{table}[t]

\centering
\label{tab:self-anotation}
\renewcommand{\arraystretch}{1.2}
\resizebox{0.38\textwidth}{!}{
\begin{tabular}{lcc}
\toprule
Training Set& GSM8K & MATH \\
\midrule
$\mathcal{D}_\textbf{DotaMathQA-w/o-aug}$  & 82.8  & 59.0 \\
$\mathcal{D}_\textbf{DotaMathQA}$ &  86.7 & 64.8\\
$\mathcal{D}_\textbf{DotaMathQA-self}$  & 81.5  & 60.1 \\
\bottomrule
\end{tabular}
}
\caption{Self-annotation vs. GPT-4 annotation.}
\end{table}

\paragraph{Analysis of Self-correction Data.}

Since we designed two types of self-correction data, we further analyze and compare them. 
Despite the 5 times difference in dataset sizes, with automatic multi-turn QA comprising 1,934 instances and rule-based multi-turn QA containing 10,150, the gains from the former on GSM8K are 0.5 lower than the latter but 0.2 lower on MATH form table \ref{gpt4vsrule}. We reduce the size of rule-based multi-turn QA to the same level of multi-turn QA, the performance of the former is worse than the latter. This suggests that self-correction data generated by GPT-4 may be more efficient than that produced by rule-based methods. Two possible reasons are: (1) The correct components within the rule-generated self-correction data might have already been learned by the model, resulting in relatively lower benefits; (2) The explanations for errors and correct answers in the rule-generated data might not fully align, leading to lower efficiency in learning to correct mistakes. However, rule-based multi-turn data requires only minimal GPT-4 annotation and avoids annotation failures. Combining the two types of self-correction data yields better performance than using either type alone, demonstrating their complementary effect.

\paragraph{Analysis of Data Augmentation Strategies.}
We compare the model fine-tuned on $\mathcal{D}_\textbf{DotaMathQA-self}$ where augmented  augmented queries are annotated using a model SFT on $\mathcal{D}_\textbf{DotaMathQA-w/o-aug}$ against DotaMath models, to investigate the benefits of varying annotation strategies. We observe that after employing self-annotation, model performance decreases on GSM8K but increases on MATH. Given the high baseline accuracy of the current model on GSM8K, self-annotation yields a performace decrease. This suggests that the effectiveness of self-annotation is related to the current model's performance on different datasets. Compared with self-annotation, as GPT-4 outperforms the current model, annotations derived from it continue to yield substantial performance improvements.

In the Appendix \ref{app: add analysis}, We also analyze the program execution simulation capabilities comparison of DotaMATH and other tool-based LLMs, the effect of filtering out buggy responses during the data augmentation phase, the impact of data augmentation and data format on different subtopics of MATH.

\section{Conclusion}
In this paper, we introduce DotaMath, a series of LLMs which adopt decomposition of thought, code assistance, intermediate process display and self-correction to solve complex math problems. To train DotaMath, we construct an instruction fine-tuning datset named DotaMathQA with 574K query-response pairs. In detail, we use query evolution to GSM8K and MATH to augment to the existing queries. Then we use gpt-4 to annotate interactive tool-use trajectories on solve the original and augmented math problems. Ultimately, we fine-tune LLaMA2, LLaMA3, LLeMA, and DeepSeekMath-Base models using DotaMathQA, resulting in the DotaMath series models. Across two in-domain and four out-of-domain mathematical benchmarks, DotaMATH achieves the best or near-best performance among all open-source models and significantly improves performance on the competition-level MATH dataset. Upon analysis, we find that our designed module provides greater assistance with difficult problems in the MATH dataset, validating the rationale of our components for complex tasks. Interestingly, our design significantly enhances the model's ability to simulate program results, allowing DotaMATH to achieve strong performance even without invoking tools. Overall, DotaMATH has further enhanced the capabilities of open-source LLMs on complex mathematical tasks, offering insights for subsequent research in LLM for mathematics.


\bibliography{anthology}















\clearpage
\appendix

\begin{center}

{\Large \textbf{Appendix}}
\end{center}

\section{Implementation Details}
We fine-tune LLaMA-2-7B, LLaMA-3-7B, llema-7B, and DeepSeekMATH-Base-7B with $D_\textbf{DotaMathQA}$ to get DotaMATH-LLMs. We train these base models with key settings including a 5e-5 learning rate, 256 global batch size, a cosine scheduler with a 3\% warm-up, a maximum context length of 4,096 (except for LLaMA2, which uses 2,048) tokens and 3 training epochs. Responses are generated using greedy decoding with a maximum sequence length of 4,096 (except for LLaMA2, which applies 2,048) and a limit of 3 maximum tool uses. Checkpoints are not selected with early stops. 
The hardware setup involves 32 NVIDIA A100 GPUs. 

\section{Additional Analysis}
\label{app: add analysis}
\subsection{Analysis on Simulating Code Execution}

\begin{table*}[ht]
\centering
\renewcommand{\arraystretch}{1.2}
\begin{tabular}{@{}lcccccc@{}}
\toprule
Model  & \multicolumn{2}{c}{GSM8K} & \multicolumn{2}{c}{MATH} \\
\cmidrule(lr){2-3} \cmidrule(lr){4-5}
         & with tool & without tool & with tool & without tool \\
\midrule
ToRA-7B     & 68.8 &  9.6($\downarrow$ 86.0\%)       & 40.1  & 7.4($\downarrow$ 81.5\%)      \\
ToRA-code-7B             & 72.6 &   11.0($\downarrow$ 84.8\%)     & 44.6  & 8.4($\downarrow$ 81.2\%)  \\
DeepSeek-MATH-Instruct     & 83.7 &  39.9($\downarrow$ 52.3\%)     & 57.4  & 24.1($\downarrow$ 58.0\%)        \\
DotaMath-deepseek            & 86.7 & 79.3 ($\downarrow$ 8.5\%)      & 64.8  & 38.2($\downarrow$ 41.0\%)  \\
\midrule
\end{tabular}
\caption{The performance comparisons of DotaMath-deepseek and other models without access to Python interpreter.}
\label{tab:simulate}
\end{table*}

To enhance the readability of the model's output, we designed a unique mechanism that enables the model's code to print sub-tasks of a problem in a chain-of-thought manner. Interestingly, this mechanism also endows the model with the capability to simulate program execution. Specifically, we enable the model to infer the results that would typically be produced by the Python interpreter, thereby deducing the answers to problems. From Table \ref{tab:simulate}, we observe that even without relying on the Python interpreter, DotaMath was still able to achieve satisfactory results, demonstrating its accurate prediction of the Python interpreter's execution outcomes. While the ToRA model and DeepseekMATH-Instruct model are also trained using the output of the Python interpreter, these models struggle to make correct inferences in its absence. On the GSM8K, DotaMath-deepseek experience only an 8.5\% decrease in accuracy without the use of tools, while both ToRA and ToRA-code see performance drops of over 80\%, and DeepSeek-MATH-Instruct suffer a 52.3\% loss. On the MATH dataset, DotaMATH experience a significant performance loss of 41.0\%, while other models showe similar proportionate declines as observed on the GSM8K. The underlying reason may be attributed to models frequently invoking Python libraries such as \textit{sympy2} for complex calculations when solving intricate problems in MATH, resulting in a significantly increased difficulty in predicting code execution. This also highlights the necessity of utilizing tools when solving complex mathematical problems.



\subsection{The Effectiveness of Data Augmentation and Format in  MATH Sub-topics}

Regarding data augmentation, we observe significant improvements in three subjects: Counting \& Prob., Geometry, and Int. Algebra, with increases over 16\%, while the smallest enhancement is seen in Algebra at 3.4\%. According to conclusions from~\citeauthor{li2023query}, the stronger the base model's capability, the higher the accuracy demand for augmented data, and the more challenging it is to enhance performance.  In Algebra, the model already reaches 78.4 without utilizing augmented data, which explains the minimal improvement. Removing decomposition of thought, except Pre-algebra, all other sub-topics are affected.  In different subjects, removing display immediate process and self-correction has the most significant impact on Pre-calculus, indicating that the harder the problem, the more necessary it is to engage in self-correction (as they are more prone to errors).

\begin{table*}[ht]
\centering
\renewcommand{\arraystretch}{1.2}
\resizebox{1.0\textwidth}{!}{
\begin{tabular}{@{}lccccccc@{}}
\toprule
Model & Algebra & Counting \& Prob. & Geometry & Int. Algebra & Number Theory & Prealgebra & Precalculus \\
\midrule
$\mathcal{D}_\textbf{DotaMathQA}$ & 81.1 ($\uparrow$ 3.4\%) & 63.5 ($\uparrow$ 17.6\%) & 47.4 ($\uparrow$ 16.5\%) & 44.0 ($\uparrow$ 18.0\%) & 76.7 ($\uparrow$ 13.8\%) & 78.3 ($\uparrow$ 8.4\%) & 46.7 ($\uparrow$ 7.1\%) \\
$\mathcal{D}_\textbf{DotaMathQA-w/o-aug}$ & 78.4 & 54.0 & 40.7 & 37.3 & 67.4 & 72.2 & 43.6 \\
$\mathcal{D}_\textbf{DotaMathQA-w/o-aug-w/o-dot}$ & 71.5 ($\downarrow$ 8.8\%) & 48.5 ($\downarrow$ 10.2\%) & 38.8 ($\downarrow$ 4.7\%) & 35.4 ($\downarrow$ 5.1\%) & 61.1 ($\downarrow$ 9.3\%) & 73.5 ($\uparrow$ 1.8\%) & 40.7 ($\downarrow$ 6.7\%) \\
$\mathcal{D}_\textbf{DotaMathQA-w/o-aug-w/o-inter}$ & 74.4 ($\downarrow$ 5.0\%) & 51.5 ($\downarrow$ 4.6\%) & 40.9 ($\uparrow$ 0.5\%) & 37.8 ($\uparrow$ 1.3\%) & 62.6 ($\downarrow$ 7.1\%) & 74.4 ($\uparrow$ 3.0\%) & 40.3 ($\downarrow$ 7.6\%) \\
$\mathcal{D}_\textbf{DotaMathQA-w/o-aug-w/o-multi}$ & 73.8 ($\downarrow$ 5.9\%) & 54.4 ($\uparrow$ 0.7\%) & 41.1 ($\uparrow$ 1.0\%) & 36.5 ($\downarrow$ 2.1\%) & 63.7 ($\downarrow$ 5.5\%) & 71.6 ($\downarrow$ 0.8\%) & 38.1 ($\downarrow$ 12.6\%) \\
\bottomrule
\end{tabular}
}
\caption{Sub-topics performance of different models on MATH.}
\label{tab:subjects}
\end{table*}

\begin{table*}[!ht]
\centering
\renewcommand{\arraystretch}{1.2}
\begin{tabular}{lcc}
\toprule
& GSM8K & MATH \\
\midrule
DotaMath-without-gpt4-anotation-with-filter &  86.7 & 64.8 \\
DotaMath-with-anotation-without-filter  & 85.5  & 63.6 \\
\bottomrule
\end{tabular}
\caption{The effectiveness of data filtering.}
\label{tab:filter}
\end{table*}

\subsection{The Effectiveness of Bug Filter}
From Table \ref{tab:filter}, we can know that without python bug filter, the accuracy of DotaMath-deepseek drops from 86.7 to 85.5 on GSM8K and from 64.8 to 63.6 on MATH.
\section{Additional results}
Our additional results are shown in Table 6 to 10.
\begin{table*}[ht]
\centering
\renewcommand{\arraystretch}{1.2}
\resizebox{0.98\textwidth}{!}{
\begin{tabular}{@{}lccccccccc@{}}
\toprule
\textbf{Model}  &\textbf{Size} &\textbf{Use Tool?} & \multicolumn{2}{c}{\textbf{In-domain}} & \multicolumn{4}{c}{\textbf{Out-of-domain}}\\
\cmidrule(lr){4-5} \cmidrule(lr){6-9} 
   &   &      & GSM8K & MATH & Mathematics & SVAMP & TabMWP & ASDiv & Average\\
\midrule
\multicolumn{8}{c}{proprietary Model} \\
\midrule

Claude-2~\citep{claude2} & - & \color{red}\ding{53}  & 85.2 &32.5 & -& -& -& -& -\\
PaLM-2~\citep{anil2023palm} &- &\color{red}\ding{53}  & 80.7 &34.3 & -& -& -& -& -\\

\bottomrule
\multicolumn{8}{c}{Open-Source Model} \\
\midrule
Qwen-1.5-110B~\citep{qwen} &110B &\color{red}\ding{53} & 85.4 &49.4 &-&86.2 &-&85.1 &-\\
LLaMA-2 RFT~\citep{yuan2023scaling} &7B & \color{red}\ding{53} & 51.2  & - & -& -& -& -& -\\
MAmmoTH-Coder~\citep{yue2023mammoth} &34B   &\color{deepgreen}\checkmark   & 72.7 &43.6 & 65.4 &85.3 & -& -& -\\
MATHCoder-CL~\citep{mathcoder}  & 34B    &\color{deepgreen}\checkmark    &81.7  &45.2        & 75.9  & -& -& - &-  \\
ToRA-CODE~\citep{ToRA}     & 34B     &\color{deepgreen}\checkmark  &80.7 &50.8 &77.9 &80.5 &70.5 &84.2 &68.7\\

WizardMATH~\citep{luo2023wizardmath}  & 70B   &\color{red}\ding{53}    &81.6  &22.7        & - & 80.0 &49.8& 76.2       
 & -\\
MAmmoTH2-8x7B-Plus~\citep{mammoth2}   &56B   &\color{red}\ding{53}   & 86.4& 47.0        & -& -& -& - & -         \\
MATHCoder-L~\citep{mathcoder}  & 70B    &\color{deepgreen}\checkmark    &83.9  &45.1        & 74.4  &84.9 &-&-&-  \\
Platypus-2~\citep{Platypus} & 70B    &\color{red}\ding{53}    &45.9  &15.0        & - & 74.3  &47.3 &72.7 & - \\
\textbf{DotaMath-LLaMA2-7B}     & 7B  &\color{deepgreen}\checkmark     & 79.6 & 50.1       & 65.0 & 84.2 &  64.1  &82.3    &64.7 \\

\textbf{DotaMath-LLaMA3-8B  }   & 8B  &\color{deepgreen}\checkmark     & 84.2 & 58.9      & 74.2 &\underline{88.3} & 70.4 & 85.1        &71.1  \\
\textbf{DotaMath-llemma-7B}     & 7B   &\color{deepgreen}\checkmark    & 81.2 & \underline{57.7}     & 76.8 & 87.8 &  77.7 & 85.1     &74.3 \\
\textbf{DotaMath-deepseek-7B}   & 7B &\color{deepgreen}\checkmark        & \underline{86.7} & \textbf{64.8} & \underline{79.1}&\textbf{89.5}&\textbf{84.2}&\underline{88.5}     &\textbf{80.1}  \\
\bottomrule
\end{tabular}
}
\caption{Additional comparison of DotaMath with open-source and proprietary source LLMs on in-domain and out-of-domain benchmarks.
}
\label{other_mainresult_table}
\end{table*}

\clearpage
\onecolumn
\section{Prompt Templates}
Our prompt templates are shown below:
\label{app:prompt}
\begin{tcolorbox}[
colback=white!10!white,
colframe=black!75!black,
title=Augmentation Prompt,
breakable]
I want you to act as a math teacher. You should think of some ways to help students do variation training for challenging competition mathematics problems. \\
Here are some ways you can refer: Introduce fractions or percentages, Combine multiple concepts, Include a conditional statement, Increase the complexity of the problem and so on.
Response with specific format, like:\\
Introduce fractions or percentages: \#\#1 new question1 \#\#1\\
Combine multiple concepts: \#\#2 new question2 \#\#2\\
...\\
Increase the complexity of the problem: \#\#10: new question10 \#\#10\\
The nth problem must be strictly limited to between \#\#n and \#\#n for our subsequent regular extraction.\\
Now you are give a math problem, think for 10 different ways.\\
Given new problem:\\
\{Query\}
\label{Augmentationprompt}
\end{tcolorbox}

\begin{tcolorbox}[
colback=white!10!white,
colframe=black!75!black,
title=System Prompt,
breakable]
<|user|>:\\
 \{Query\}\\
 <|assistant|>:\\
\label{systemprompt}
\end{tcolorbox}

\begin{tcolorbox}[
colback=white!10!white,
colframe=black!75!black,
title=Generative Prompt,
breakable]
You are an exceptionally strong competitor in both math and programming contests, proficient in a wide range of mathematical knowledge and skilled in Python programming. Your command of Pre-algebra, Algebra, Number Theory, Counting and Probability, Geometry, Intermediate Algebra, and Precalculus is unparalleled. Your thinking is meticulous and profound, and the code you write always runs flawlessly and without error.

Integrate step-by-step reasoning and Python code to solve math problems using the following guidelines:\\
\begin{enumerate}
    \item Break the problem into subtasks.
    \item Write functions to solve the problem; the function should not take any arguments.
    \item Print the results of every subtask in the Python code, using the intermediate variables in Python programs to represent intermediate results, refer to the example below.
    \item When writing the python program, avoid using decimal. Utilize functions from \texttt{sympy} and other necessary Python library, and simplify all fractions and square roots without converting them to decimal values.
    \item Print the final answer on the last line.
\end{enumerate}

Here is an example you may refer to:

\textit{Problem:} Let 
\[f(x) = \left\{
\begin{array}{cl}
    ax+3, & \text{if } x>2, \\
    x-5, & \text{if } -2 \leq x \leq 2, \\
    2x-b, & \text{if } x <-2.
\end{array}
\right.\]
Find $a+b$ if the piecewise function is continuous (which means that its graph can be drawn without lifting your pencil from the paper).

\textit{Solution:}
We can decompose this problem into following sub-tasks:
\begin{enumerate}
    \item Solve for $a$ by equating $ax + 3$ to $x - 5$ at $x = 2$.
    \item Solve for $b$ by equating $x - 5$ to $2x - b$ at $x = -2$.
    \item Add the values of $a$ and $b$ together to find the sum.
\end{enumerate}
\verb|```python|
\begin{verbatim}
from sympy import symbols, Eq, solve

def sum_a_and_b():
    a = symbols('a')
    b = symbols('b')
    equation1 = Eq(a * 2 + 3, 2 - 5)
    equation2 = Eq(-2 - 5, 2*(-2) - b)
    solution_a = solve(equation1, a)
    solution_b = solve(equation2, b)
    sum_ab = solution_a[0] + solution_b[0]
    # print the results of every subtask
    print(f"Equating the function at x = 2 gives us the equation {equation1}.")
    print(f"Solving this equation gives us the value of a: a = {solution_a[0]}.")
    print(f"Equating the function at x = -2 gives us the equation {equation2}.")
    print(f"Solving this equation gives us the value of b: b = {solution_b[0]}.")
    print(f"hence, a + b equals to {solution_a[0]}+{solution_b[0]} = {sum_ab}.")
    return sum_ab

sum_ab = sum_a_and_b()
# print the final answer
print(sum_ab)
\end{verbatim}
\verb|```|
\verb|```output|
\begin{verbatim}
Output:
Equating the function at x = 2 gives us the equation Eq(2*a + 3, -3).
Solving this equation gives us the value of a: a = -3.
Equating the function at x = -2 gives us the equation Eq(-7, -b - 4).
Solving this equation gives us the value of b: b = 3.
hence, a + b equals to -3+3 = 0.
0
\end{verbatim}
We find that the sum of $a$ and $b$ is $0$. This ensures the piecewise function is continuous across its entire domain. Therefore, the final answer is $\boxed{0}$.

\label{generativeprompt}
\end{tcolorbox}

\begin{tcolorbox}[
colback=white!10!white,
colframe=black!75!black,
title=Correction Prompt,
breakable]
You are an exceptionally strong competitor in both math and programming contests, proficient in a wide range of mathematical knowledge and skilled in Python programming. Your command of Pre-algebra, Algebra, Number Theory, Counting and Probability, Geometry, Intermediate Algebra, and Precalculus is unparalleled. Your thinking is meticulous and profound, and the code you write always runs flawlessly and without error. You solve the problem with PLAN-CODE format.

Integrate step-by-step reasoning and Python code to solve math problems using the following guidelines:
\begin{enumerate}
    \item Break the problem into subtasks.
    \item Write functions to solve the problem; the function should not take any arguments.
    \item Print the results of every subtask in the Python code.
    \item When writing the python program, avoid using decimal. Utilize functions from sympy and other necessary python library, and simplify all fractions and square roots without converting them to decimal values.
    \item Print the numeric answer on the last line.
\end{enumerate}

Here is an example you may refer to:

\textbf{Problem:}
Let 
\[
f(x) = \left\{
\begin{array}{cl}
    ax+3, & \text{if } x>2, \\
    x-5, & \text{if } -2 \leq x \leq 2, \\
    2x-b, & \text{if } x < -2.
\end{array}
\right.
\]
Find $a+b$ if the piecewise function is continuous (which means that its graph can be drawn without lifting your pencil from the paper).

\textbf{Solution:}
We can decompose this problem into the following sub-tasks:
\begin{enumerate}
    \item Solve for $a$ by equating $ax + 3$ to $x - 5$ at $x = 2$.
    \item Solve for $b$ by equating $x - 5$ to $2x - b$ at $x = -2$.
    \item Add the values of $a$ and $b$ together to find the sum.
\end{enumerate}

\verb|```python|
\begin{verbatim}
from sympy import symbols, Eq, solve

def sum_a_and_b():
    a = symbols('a')
    b = symbols('b')
    equation1 = Eq(a * 2 + 3, 2 - 5)
    equation2 = Eq(-2 - 5, 2*(-2) - b)
    solution_a = solve(equation1, a)
    solution_b = solve(equation2, b)
    sum_ab = solution_a[0] + solution_b[0]
    # print the results of every subtask
    print(f"Equating the function at x = 2 gives us the equation {equation1}.")
    print(f"Solving this equation gives us the value of a: a = {solution_a[0]}.")
    print(f"Equating the function at x = -2 gives us the equation {equation2}.")
    print(f"Solving this equation gives us the value of b: b = {solution_b[0]}.")
    print(f"hence, a + b equals to {solution_a[0]}+{solution_b[0]} = {sum_ab}.")
    return sum_ab

sum_ab = sum_a_and_b()
# print the final answer
print(sum_ab)
\end{verbatim}
\verb|```|

Given a math problem, an incorrect solution, and you need to correct the incorrect solution. The response format is like: (short error interpretations)+(new solution).
\label{Correctionprompt}
\end{tcolorbox}

\begin{tcolorbox}[
colback=white!10!white,
colframe=black!75!black,
title=Explanatory Prompt,
breakable]
You are an exceptionally strong competitor in both math and programming contests, proficient in a wide range of mathematical knowledge and skilled in Python programming. Your command of Pre-algebra, Algebra, Number Theory, Counting and Probability, Geometry, Intermediate Algebra, and Precalculus is unparalleled. Your thinking is meticulous and profound, and the code you write always runs flawlessly and without error.\\
Given a math problem, an incorrect solution and a correct solution, your task is to give the incorrect reason for the incorrect solution. Note that the reason is not allowed to be too long. You don't need to give the correct answer. You need to start your answer with: 'The solution is wrong since...'
\label{Explanatoryprompt}
\end{tcolorbox}
\twocolumn




\onecolumn
\section{Case Study}
\subsection{Singl-turn Successful cases}
\begin{tcolorbox}[
colback=white!10!white,
colframe=black!75!black,
title=Single-trun successful case in GSM8K,
breakable]
\textit{Problem:} Janet’s ducks lay 16 eggs per day. She eats three for breakfast every morning and bakes muffins for her friends every day with four. She sells the remainder at the farmers' market daily for 2 per fresh duck egg. How much in dollars does she make every day at the farmers' market?

\textit{Solution:}
To solve the problem of calculating Janet's daily earnings from selling fresh duck eggs at the farmers' market, we can follow these steps:
\begin{enumerate}
\item Calculate the total number of eggs laid by the ducks each day.
\item Subtract the number of eggs Janet eats for breakfast and the number used in baking muffins to find the number of eggs available for sale.
\item Calculate the earnings from selling the remaining eggs at 2 per egg.
\end{enumerate}
\verb|```python|
\begin{verbatim}
from sympy import symbols, Eq, solve

def calculate_daily_earnings():
# Define the variables
total_eggs_per_day = 16
eggs_for_breakfast = 3
eggs_for_muffins = 4
price_per_egg = 2
# Calculate the number of eggs available for sale
eggs_for_sale = total_eggs_per_day - (eggs_for_breakfast + eggs_for_muffins)

# Calculate the earnings from selling the eggs
daily_earnings = eggs_for_sale * price_per_egg

# Print the results of every subtask
print(f"Total number of eggs laid per day: {total_eggs_per_day}")
print(f"Number of eggs eaten for breakfast: {eggs_for_breakfast}")
print(f"Number of eggs used in baking muffins: {eggs_for_muffins}")
print(f"Number of eggs available for sale: {eggs_for_sale}")
print(f"Daily earnings from selling eggs: ${daily_earnings}")

return daily_earnings
Calculate and print the daily earnings
daily_earnings = calculate_daily_earnings()
print(daily_earnings)
\end{verbatim}
\verb|| \verb|output|
\begin{verbatim}
Output:
Total number of eggs laid per day: 16
Number of eggs eaten for breakfast: 3
Number of eggs used in baking muffins: 4
Number of eggs available for sale: 9
Daily earnings from selling eggs: $18
18
\end{verbatim}
Following these calculations, Janet makes $\boxed{18}$ dollars every day at the farmers' market by selling fresh duck eggs.
\end{tcolorbox}

\begin{tcolorbox}[
colback=white!10!white,
colframe=black!75!black,
title=Single-trun successful case in MATH,
breakable]
\textit{Problem:} Find all positive integer values of $c$ such that the equation $x^2-7x+c=0$ only has roots that are real and rational. Express them in decreasing order, separated by commas.

\textit{Solution:}
For the roots to be real and rational, the discriminant must be a perfect square. Therefore, $(-7)^2-4 \cdot 1 \cdot c = 49-4c$ must be a perfect square. The only positive perfect squares less than 49 are $1$, $4$, $9$, $16$, $25$, and $36$. The perfect squares that give an integer value of $c$ are $1$, $9$, and $25$. Thus, we follow these steps:
\begin{enumerate}
\item Solve the equation $49-4c=1$ to find a value of $c$.
\item Solve the equation $49-4c=9$ to find another value of $c$.
\item Solve the equation $49-4c=25$ to find the last value of $c$.
\end{enumerate}
\verb|```python|
\begin{verbatim}
from sympy import symbols, solve, sqrt

def find_c_values():
c = symbols('c')
discriminant = (-7)**2 - 41c # Compute the discriminant
# List to store valid c values
valid_c_values = []

# Check for each c if the discriminant is a perfect square
for c_value in range(1, 100):  # Assuming a reasonable upper limit for c
    disc_value = discriminant.subs(c, c_value)
    if sqrt(disc_value).is_integer():
        valid_c_values.append(c_value)
        print(f"For c = {c_value}, the discriminant is {disc_value}.")

# Print the valid c values in decreasing order
valid_c_values.sort(reverse=True)
print("Valid c values in decreasing order:", ", ".join(map(str, valid_c_values)))

return valid_c_values
valid_c_values = find_c_values()
\end{verbatim}
\begin{verbatim}
Output:
For c = 6, the discriminant is 25.
For c = 10, the discriminant is 9
For c = 12, the discriminant is 1.
Valid c values in decreasing order: 12, 10, 6
\end{verbatim}
Following these steps, the positive integer values of $c$ that ensure the equation $x^2 - 7x + c=0$ has real and rational roots by making the discriminant a perfect square, are $12$, $10$, and $6$. Therefore, the final answer is $\boxed{12, 10, 6}$.
\end{tcolorbox}

\subsection{Multi-turn Successful cases}
\begin{tcolorbox}[
colback=white!10!white,
colframe=black!75!black,
title=Multi-turn successful case in GSM8K,
breakable]
\textit{Problem:}India has 4 more than half the number of states in the USA. What's the total number of states in both countries together?
\textit{Solution}:To solve the problem, we can break it down into the following subtasks:

\begin{enumerate}
\item Define the number of states in the USA.
\item Calculate the number of states in India based on the given relationship.
\item Sum the number of states in both countries.
\end{enumerate}

Let's implement this in Python using the sympy library to handle symbolic computation and ensure that all operations are exact.

\verb|```python|
\begin{verbatim}
from sympy import symbols, Eq, solve

def total_states():
# Define the number of states in the USA
states_usa = symbols('states_usa')
# Given that India has 4 more than half the number of states in the USA
states_india = Eq(states_india, 4 + 1/2 * states_usa)

# Solve for the number of states in the USA (assuming the number of states 
# in the USA is known) For the sake of example, let's assume the USA has 
# 50 states
states_usa_value = 50
states_india_value = solve(states_india.subs(states_usa, states_usa_value))[0]

# Calculate the total number of states
total = states_usa_value + states_india_value

# Print the results of every subtask
print(f"Number of states in the USA: {states_usa_value}")
print(f"Number of states in India (4 more than half of the USA): 
      {states_india_value}")
print(f"Total number of states in both countries: {total}")

return total
Call the function and print the final answer
total_number_of_states = total_states()
print(total_number_of_states)
\end{verbatim}
\verb|| \verb|output|
\begin{verbatim}
UnboundLocalError: local variable 'states_india' referenced before assignment
\end{verbatim}
The error occurs because the variable states india is referenced before it is assigned a value in the equation. This is a typical scenario where the variable should be defined before it is used in an equation. Let's correct this by defining states india as a symbol and then using it in the equation.
\verb|```python|
\begin{verbatim}
from sympy import symbols, Eq, solve

def total_states():
# Define the number of states in the USA
states_usa = symbols('states_usa')
# Define the number of states in India as a symbol
states_india = symbols('states_india')

# Given that India has 4 more than half the number of states in the USA
equation = Eq(states_india, 4 + 1/2 * states_usa)

# Solve for the number of states in the USA (assuming the number of states
# in the USA is known) For the sake of example, let's assume the USA has 50 states
states_usa_value = 50
states_india_value = solve(equation.subs(states_usa, states_usa_value))[0]

# Calculate the total number of states
total = states_usa_value + states_india_value

# Print the results of every subtask
print(f"Number of states in the USA: {states_usa_value}")
print(f"Number of states in India (4 more than half of the USA): 
      {states_india_value}")
print(f"Total number of states in both countries: {total}")

return total
Call the function and print the final answer
total_number_of_states = total_states()
print(total_number_of_states)
\end{verbatim}
\verb|| \verb|output|
\begin{verbatim}
Number of states in the USA: 50
Number of states in India (4 more than half of the USA): 29.0000000000000
Total number of states in both countries: 79.0000000000000
79.0000000000000
\end{verbatim}
The total number of states in both the USA and India, calculated by adding the number of states in the USA (50) to the number of states in India (29), is $\boxed{79}$.
\end{tcolorbox}

\begin{tcolorbox}[
colback=white,
colframe=black,
title=Multi-turn successful case in MATH,
breakable
]
\textit{Problem:}
A lattice point in the $xy$-plane is a point both of whose coordinates are integers (not necessarily positive).  How many lattice points lie on the hyperbola $x^2-y^2=17$?\\
\textit{Solution:}
To find the number of lattice points on the hyperbola $x^2 - y^2 = 17$, we can follow these steps:
\begin{enumerate}
    \item Recognize that $x^2 - y^2 = (x + y)(x - y)$.
    \item Since $x$ and $y$ are integers, $(x + y)$ and $(x - y)$ must be factors of $17$.
    \item The number $17$ is a prime number, so its only factors are $1$ and $17$ or $-1$ and $-17$.
    \item We will set up equations for $x + y = 17$ and $x - y = 1$, and also for $x + y = -17$ and $x - y = -1$.
    \item Solve these equations to find the integer solutions for $x$ and $y$.
    \item Count the number of solutions to determine the number of lattice points on the hyperbola.
\end{enumerate}

\begin{verbatim}
from sympy import symbols, Eq, solve

# Define symbolic variables
x, y = symbols('x y')

# Define the equations based on the factors of 17
equations = [
    (Eq(x + y, 17), Eq(x - y, 1)),
    (Eq(x + y, -17), Eq(x - y, -1))
]

# Function to solve the equations and count the lattice points
def count_lattice_points():
    lattice_points = []
    for eq1, eq2 in equations:
        solutions = solve((eq1, eq2), (x, y))
        lattice_points.append(solutions)
        print(f"Solving the equations {eq1} and {eq2} result: {solutions}.")
    return len(lattice_points)

# Count the lattice points on the hyperbola
number_of_lattice_points = count_lattice_points()
print(number_of_lattice_points)
\end{verbatim}

\textbf{Output:}
\begin{verbatim}
Solving the equations Eq(x + y, 17) and Eq(x - y, 1) result: {x: 9, y: 8}.
Solving the equations Eq(x + y, -17) and Eq(x - y, -1) result: {x: -9, y: -8}.
2
\end{verbatim}
The solution is wrong since it fails to consider the negative factors of 17, which are also necessary to find all lattice points on the hyperbola. The correct approach should include both positive and negative factors of 17, leading to additional solutions. Let's correct the solution.
To find the number of lattice points on the hyperbola $x^2 - y^2 = 17$, we can follow these steps:
\begin{enumerate}
    \item Recognize that $x^2 - y^2 = (x + y)(x - y)$.
    \item Since $x$ and $y$ are integers, $(x + y)$ and $(x - y)$ must be factors of $17$.
    \item List all pairs of factors of $17$, which are $(1, 17)$ and $(-1, -17)$, and their reverses.
    \item For each pair of factors, solve the system of equations $(x + y = a)$ and $(x - y = b)$ where $(a, b)$ is a pair of factors.
    \item The solutions to the system of equations will give us the lattice points.
    \item Count the number of unique lattice points obtained from the solutions.
\end{enumerate}
\begin{verbatim}
from sympy import symbols, Eq, solve

# Define symbolic variables
x, y = symbols('x y')

# Define the factors of 17
factors = [(1, 17), (-1, -17), (17, 1), (-17, -1)]

# Function to find lattice points
def find_lattice_points(factors):
    lattice_points = []
    for a, b in factors:
        # System of equations based on the factors
        eq1 = Eq(x + y, a)
        eq2 = Eq(x - y, b)
        # Solve the system of equations
        solution = solve((eq1, eq2), (x, y))
        # Add the solution to the list of lattice points
        lattice_points.append(solution)
    return lattice_points

# Find the lattice points
lattice_points = find_lattice_points(factors)

# Print the detailed reasoning process
print(f"The factors of 17 are: {factors}")
print(f"The lattice points on the hyperbola are: {lattice_points}")

# Print the final answer
print(len(lattice_points))
\end{verbatim}

\textbf{Output:}
\begin{verbatim}
The factors of 17 are: [(1, 17), (-1, -17), (17, 1), (-17, -1)]
The lattice points on the hyperbola are: [{x: 9, y: -8}, {x: -9, y: 8}, 
{x: 9, y: 8}, {x: -9, y: -8}]
4
\end{verbatim}

The code successfully finds all lattice points on the hyperbola $x^2 - y^2 = 17$ by considering all pairs of factors of $17$ and solving the system of equations for each pair. Since the hyperbola equation can be factored into $(x + y)(x - y) = 17$, and $17$ is a prime number, it only has the positive factors $(1, 17)$ and $(17, 1)$, and their negative counterparts $(-1, -17)$ and $(-17, -1)$. Solving these pairs as systems of equations yields the lattice points $(9, -8)$, $(-9, 8)$, $(9, 8)$, and $(-9, -8)$. Therefore, there are $\boxed{4}$ lattice points on the hyperbola.
\end{tcolorbox}

\newpage
\end{document}

%% file: latex/intro.tex
\section{Introduction}

The emergence of large language models (LLMs) ~\citep{ouyang2022training,anil2023palm,openai2024gpt4,gemini,claude-3,qwen2} has profoundly revolutionized a diverse range of natural language processing benchmarks~\citep{chen2021evaluating,longpre2023flan,wei2023chainofthought,luo2023wizardcoder,qiao2024we,song2024cs}.
However, in the challenging field of mathematical reasoning, enabling open-source LLMs to possess reasoning abilities for complex mathematical tasks remains a significant challenge~\citep{ToRA,yue2023mammoth,yue2024mammoth2scalinginstructionsweb}.

Existing works have attempted to enhance the reasoning capabilities of LLMs through methods such as chain-of-thought (\textbf{COT})~\citep{cot}, program-of-thought (\textbf{POT})~\citep{pot,pal}, and tool-integrated reasoning approaches (\textbf{Tool-based})~\citep{ToRA,mathcoder}.
The Tool-based approach effectively merges COT's semantic and abstract reasoning with POT's computational precision, demonstrating commendable performance. Meanwhile, 
several efforts utilize state-of-the-art proprietary models like GPT-4 to augment existing mathematical reasoning datasets~\citep{yu2023metamath,luo2023wizardmath,li2023query}, thereby improving the reasoning capabilities of LLMs during the supervised fine-tuning (SFT) phase. Building on prior works, open-source LLMs have achieved commendable performance on simple math problems. For example, on GSM8K~\citep{gsm8k} which contains grade school math word problems, many math-specific LLMs exceeding 80\% accuracy. However, they continue to struggle with complex mathematical reasoning tasks. For instance, on MATH~\citep{MATH} dataset comprising challenging competition problems, almost all open-source LLMs cannot exceed 60\% accuracy. Through our investigation, we find that these open-source LLMs lack meticulous design for complex mathematical tasks. They do not consider the necessity of task decomposition for complex tasks, nor do they account for the need for LLMs to obtain more feedback signals from tools to facilitate comprehensive analysis.

\begin{figure*}[ht]
    \centering
    \includegraphics[width=0.98\linewidth]{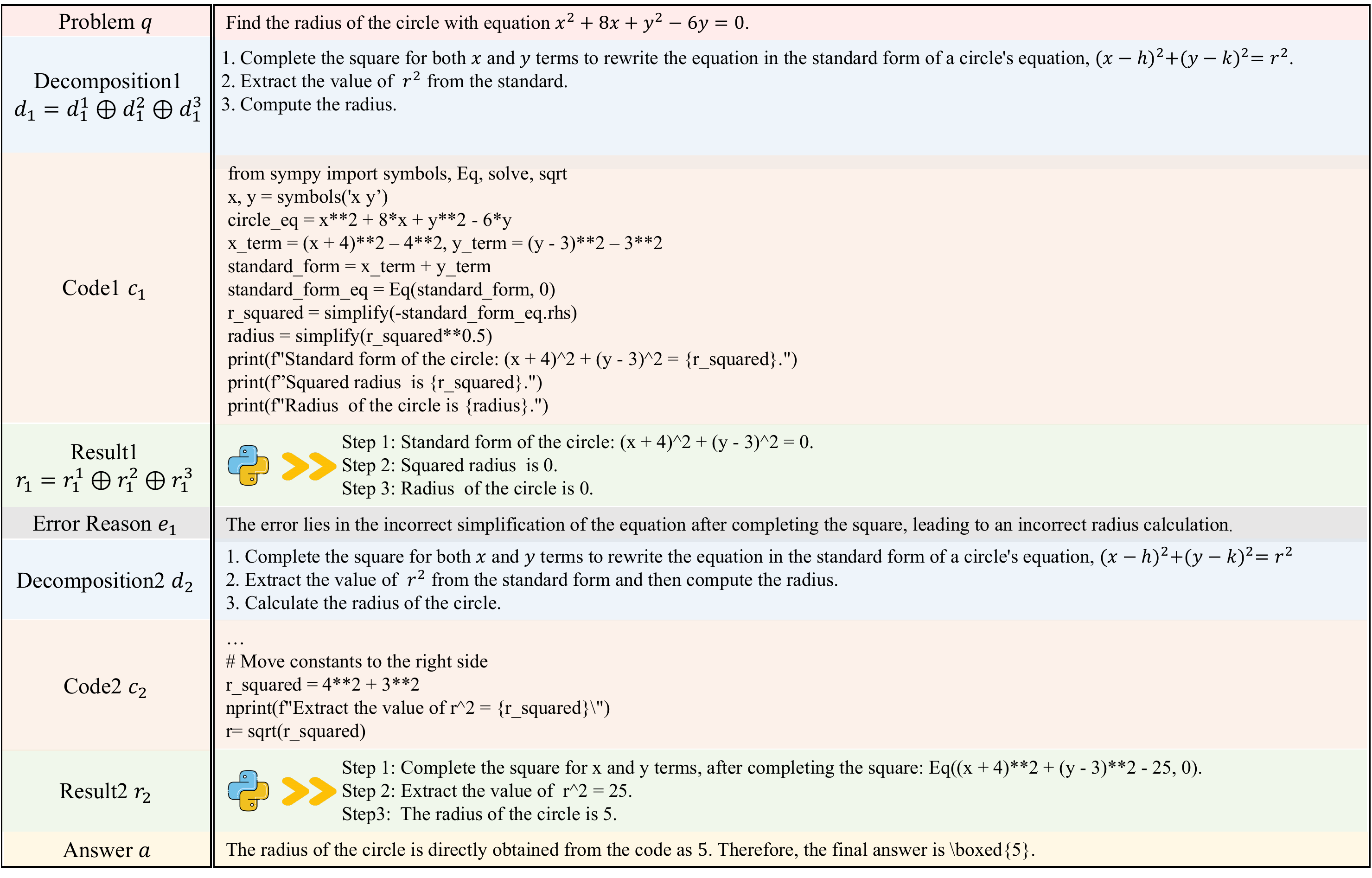}
    \caption{An example shows how the DotaMath use decomposition and self-correction to solve a probem form MATH test set. For problem $p$, the model initially decomposes it into three subtasks and subsequently writes corresponding code. The Python interpreter then executes this code, producing three subresults. Upon identifying that the results are incorrect, the model elucidates the error's cause and undertakes a revised decomposition. Following this, new code is crafted and executed by the Python interpreter, leading to the correct resolution of the problem and presentation of the final solution.}\vspace{-5pt}
    \label{data_format}
\end{figure*}

To improve the capabilities of open-source LLMs in complex mathematical reasoning tasks, this paper introduces DotaMath models, a series of LLMs which employ the \textbf{D}ecomposition \textbf{o}f \textbf{t}hought with code \textbf{a}ssistance and self-correction for \textbf{math}ematical Reasoning. There are three special designs in DotaMath for complex mathematical tasks, as depicted in Figure~\ref{data_format}. 
(1) \textbf{Decomposition of thought}: The principle of divide-and-conquer often allows complex tasks to be decomposed into more easily solvable subtasks. Inspired by some relevant works~\citep{khot2023decomposed,motcoder}, DotaMath break down mathematical problems into logical subtasks and use code to solve them.
(2) \textbf{Intermediate process display}: While previous tool-based math-specifical LLMs~\citep{mathcoder,ToRA,deepseek} obtain only single mathematical expressions from code interpreters, we aim for DotaMATH to receive more fine-grained feedback from the code interpreter for subsequent comprehensive analysis. 
To achieve this goal, we facilitate the model to print the results of all subtasks in the form of chain of thought  within the code.  This design also contributes to enhancing the human readability of the model's responses.
(3) \textbf{Self-correction}: When solving complex tasks, the issue of not succeeding at once sometimes arises. Self-reflection and correction are appropriate for resolving this problem~\citep{Reflexion,reflection2}. We construct two types of instruction fine-tuning data to endow DotaMath with self-correction capabilities.

For data construction, we introduce an instruction-following dataset named \textbf{DotaMathQA}, based on the human-curated mathematical datasets GSM8K and MATH. As shown in Figure \ref{data_produce}, DotaMathQA contains two types of data: one involves data that requires a single invocation of Python code, referred to as \textbf{Single-turn QA}; the other includes data with a self-correction process, necessitating multiple invocations of Python code, referred to as \textbf{Multi-turn QA}. Inspired by previous works~\citep{luo2023wizardmath,yu2023metamath,li2023query}, we adopt the query evolution to bootstrap mathematical questions in GSM8K and MATH with the augmentation prompt in Appendix \ref{Augmentationprompt}.

With DotaMathQA, we fine-tune a series of backbone models, including Llama2-7B~\citep{llama2}, Llama3-8B~\citep{llama3}, Llemma-7B~\citep{llemma} and DeepSeekMath-Base-7B~\citep{deepseek}. As shown in Table \ref{mainresult_table}, 
DotaMath outperforms open-source models across all scales on in-domain GSM8K and MATH datasets and four out-of-domain mathemathical benchmarks overall. Notably, DotaMath-deepseek-7B showcases an outstanding performance of 64.8\% on the competitive MATH dataset. Besides, DotaMath-deepseek-7B maintains strong competitiveness on GSM8K (86.7\%) and achieves an anverage of 80.1\% on two in-domain benchmarks and four out-of-domain benchmarks. Looking ahead, we believe the DotaMath paradigm will pave a new avenue for solving complex mathematical tasks.

